\documentclass[fleqn]{2022AMSE}
\usepackage{graphicx}
\usepackage{subcaption}
\usepackage{booktabs}
\usepackage{enumitem} 
\usepackage{setspace}

\usepackage{bm}
\usepackage{float}
\usepackage{multirow}

\usepackage[backend=biber, 
            style=numeric-comp,
            sorting=none,
            giveninits=true,
            maxbibnames=20,
            natbib=true]{biblatex}
\hypersetup{
    colorlinks=true,
    linkcolor=blue,
    urlcolor=red,
    citecolor=green
}

\begin{document}

\ensubject{Fluid Dynamics}

\ArticleType{RESEARCH PAPER}
\Year{2025}
\Vol{41}
\DOI{10.1007/s10409-025-25314-x}
\ArtNo{125314}
\ReceiveDate{13 April 2025}
\AcceptDate{30 May 2025}
\OnlineDate{???}

\title{Event-based multi-view photogrammetry for high-dynamic, high-velocity target measurement}{Event-based multi-view photogrammetry for high-dynamic, high-velocity target measurement}

\author[1,2]{Lei Taihang}{}%
\author[1,2]{Guan Banglei}{guanbanglei12@nudt.edu.cn}
\author[3]{Liang Minzu}{} 
\author[3]{Li Xiangyu}{} 
\author[1,2]{\\Liu Jianbing}{}
\author[1,2]{Tao Jing}{}
\author[1,2]{Shang Yang}{}
\author[1,2]{Yu Qifeng}{}

\AuthorMark{T. Lei}

\AuthorCitation{T. Lei, B. Guan, M. Liang, X. Li, J. Liu, J. Tao, Y. Shang, and Q. Yu}

\address[1]{The College of Aerospace Science and Engineering, National University of Defense Technology, 109 Deya Rd, \\Changsha, Hunan, China}
\address[2]{The Hunan Provincial Key Laboratory of Image Measurement and Vision Navigation, 109 Deya Rd, \\Changsha, Hunan, China}
\address[3]{The College of Science, National University of Defense Technology, 109 Deya Rd, Changsha, Hunan, China}

\contributions{Executive Editor: ???}

\abstract{The characterization of mechanical properties for high-dynamic, high-velocity target motion is essential in industries. It provides crucial data for validating weapon systems and precision manufacturing processes etc. However, existing measurement methods face challenges such as limited dynamic range, discontinuous observations, and high costs. This paper presents a new approach leveraging an event-based multi-view photogrammetric system, which aims to address the aforementioned challenges. First, the monotonicity in the spatiotemporal distribution of events is leveraged to extract the target's leading-edge features, eliminating the tailing effect that complicates motion measurements. Then, reprojection error is used to associate events with the target's trajectory, providing more data than traditional intersection methods. Finally, a target velocity decay model is employed to fit the data, enabling accurate motion measurements via ours multi-view data joint computation. In a light gas gun fragment test, the proposed method showed a measurement deviation of $4.47\%$ compared to the electromagnetic speedometer.

\textbf{Statement: The final publication is available at link.springer.com via DOI: 10.1007/s10409-025-25314-x}}


\setlength{\textheight}{23.6cm}
\thispagestyle{empty}

\maketitle
\setlength{\parindent}{1em}

\vspace{-1mm}
\begin{multicols}{2}

\section{Introduction}
\noindent The characterization of mechanical properties for high-dynamic, high-velocity target motion is critical in equipment development and other industry fields\cite{1}. It directly affects tasks like dynamic load spectrum accuracy, aerodynamic shape optimization, and the quantification of structural failure mechanisms under extreme 
\vspace{1em}
\rule[-10pt]{3.5cm}{0.05em} 
\begin{spacing}{1.32}
\noindent\scriptsize{*Corresponding author. E-mail address:}\\
\scriptsize{guanbanglei12@nudt.edu.cn (Banglei Guan)} \\
\scriptsize{Executive Editor: ???} \\
\end{spacing}
\noindent conditions, etc.\cite{2}. This tasks provides essential data for key engineering challenges, such as thermal protection system design for high-velocity aircraft, warhead damage prediction, and reliability verification of precision manufacturing processes. However, current measurement methods face limitations due to the target's high-velocity, small size, and interference from fire and light\cite{3}. These challenges result in constraints on dynamic range, insufficient time resolution, and significant costs when obtaining mechanical parameters like initial velocity and velocity decay\cite{4}\cite{5}. Accurately and efficiently capturing motion data for high-dynamic, high-velocity targets remains a major mechanical challenge\cite{6}.

In recent years, the rapid development of neuromorphic vision (a bio-inspired vision technology)\cite{7} has provided an opportunity to address the challenge of measuring such targets. These cameras simulate the visual signal processing mechanisms of the human brain\cite{8}, selectively perceiving only the regions of the scene that change. This approach significantly reduces data volume and cost. Event cameras, as the most prominent type of neuromorphic vision camera\cite{9}\cite{10}, offer advantages such as a high dynamic range ($140$dB), low latency ($1\upmu\mathrm{s}$), and low cost (with spatiotemporal resolution pricing about $10\%$ that of high-speed cameras)\cite{11}. These features make them particularly well-suited for observing high-dynamic, high-velocity targets\cite{12}. However, due to the asynchronous nature of event triggering, the traditional image frame processing paradigm is not directly applicable to event streams\cite{13}\cite{14}.

This paper proposes an innovative approach to measure the motion parameters of high-dynamic, high-velocity targets by constructing a multi-view photogrammetric system using multiple event cameras. First, the monotonicity in the spatiotemporal distribution of events is utilized to extract the target's leading-edge features, mitigating the influence of tailing effects\cite{15} on motion measurements. Second, the reprojection error is used to guide the association between events and 3D points of the target's trajectory. Finally, a target velocity decay theoretical model is employed to fit the data obtained from the event stream, enabling the measurement of motion parameters. The key contributions of this paper are as follows:

\begin{itemize}
\item The monotonicity in the spatiotemporal distribution of events is leveraged to extract the target's leading-edge features, effectively mitigating the influence of tailing effects on motion measurements for high-dynamic, high-velocity targets.
\item The use of reprojection error is introduced to guide the association between events and the 3D target's trajectory, providing more data than traditional intersection methods.
\item A target velocity decay model is employed to fit the data from the event stream, enabling accurate motion measurements via ours multi-view data joint computation.
\end{itemize}

\section{Related work}
\noindent The methods for high-dynamic, high-velocity targets' motion measurements can be broadly categorized into contact-based and non-contact methods\cite{16}\cite{17}\cite{18}. 

Contact-based methods, such as the witness plate method, require placing multiple witness plates along the target's trajectory\cite{19}. These plates capture the target's motion, using prior information like sensor or camera data and plate spacing to measure its parameters\cite{20}. This method is cost-effective but requires reconfiguring the plates for each test. Additionally, it is a damped test, which may interfere with the target's motion and introduce measurement errors. 

Non-contact methods include the light screen velocimetry, the electrical measurement, and high-speed photography. The light screen velocimetry improves upon the witness plate method by offering a non-damped test with higher velocity measurement accuracy. However, it can only measure the target's velocity when it passes through the light curtain and does not allow for continuous observation\cite{21}\cite{22}. The electrical measurement uses electromagnetic sensors to detect changes in the electromagnetic field caused by the target's motion. This allows for the evaluating of the target's velocity, but it is susceptible to electromagnetic interference and cannot measure non-magnetized targets, such as tungsten core projectiles\cite{23}. High-speed photography captures real-time high-speed images of the target and uses optical imaging geometry to measure its motion parameters\cite{24}\cite{25}. This method provides non-contact, continuous observation with retrievable data and high flexibility\cite{26}. However, high-speed cameras face challenges related to the trade-offs between dynamic range, spatial resolution, and temporal resolution\cite{27}. Issues such as overexposure and potential overheating can also impact the accuracy and precision of data collection. 

In conclusion, current methods for measuring the motion parameters of high-dynamic, high-velocity targets need improvements in dynamic range, spatiotemporal resolution, flexibility, and cost-effectiveness.

\begin{figure*}
  \centering
  \includegraphics[width=0.93\textwidth,height=!]{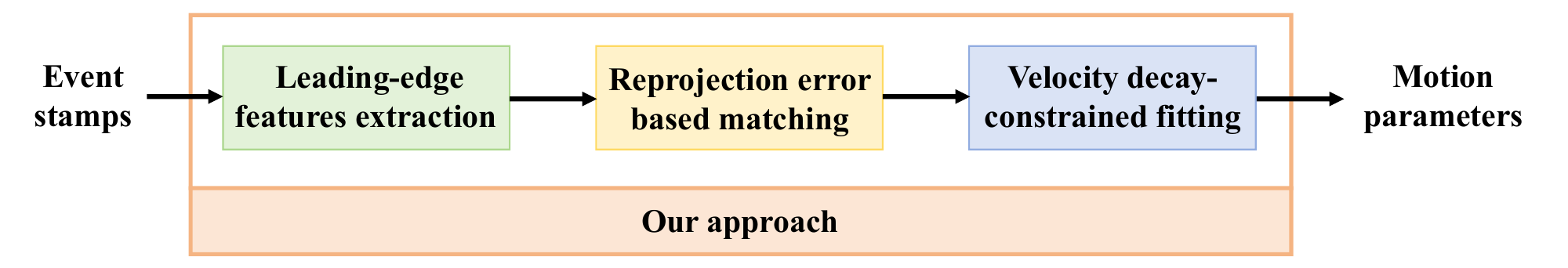}
  \caption{Overview of our proposed method. Our method directly processes asynchronous event data without accumulating event frames.}
\end{figure*}

\section{Method}
\noindent This section introduces the fundamental principles of high-dynamic, high-velocity target measurement using an event-based multi-view photogrammetric system. The basic approach is illustrated in Fig. 1. First, the basic imaging principles of event cameras are explained. Then, the impact of tailing effect on high-dynamic, high-velocity target measurement is analyzed. A method for leading-edge features extraction is proposed, utilizing the monotonicity of the spatiotemporal distribution of events. Next, the event-triggering uncertainty in different views is analyzed. Reprojection error is proposed as a criterion to associate events with the target's motion trajectory. Finally, a velocity decay model is employed to fit the data obtained from the event stream. The motion parameters of high-dynamic, high-velocity targets are measured accurately via ours multi-view data joint computation.

\subsection{Imaging principles of event cameras}
\noindent An event camera is a novel bio-inspired visual sensor. Its operating principle is based on logarithmic operations. It converts light intensity signals into voltage signals and continuously monitors the voltage changes of each pixel. This process can be represented by Eq. (1):
\begin{eqnarray}
    \left\{
        \begin{array}{l}
            \Delta L=L(x, y, t)-L(x, y, t-\delta)\\
            \Delta V=log \lvert \Delta L \rvert
        \end{array}
    \right.
\end{eqnarray}
where $\Delta L$ represents the change in light intensity, $\Delta V$ represents the change in voltage, $(x,y)$ denotes the pixel coordinates, $t$ represents time, and $\delta$ is the infinitesimal time interval (around $1\upmu\mathrm{s}$). When the voltage change of a pixel $\Delta V$ exceeds a predefined threshold $\Phi$, the event camera outputs an event $e(x, y, p, t)$ determined by the direction of the light intensity change. Eq. (2) describes the principle of event polarity $p$ assignment:
	\begin{eqnarray}
            p=\frac{\Delta L}{\lvert \Delta L \rvert}, if \quad \Delta V \ge \Phi.
    \end{eqnarray}
Each pixel of the event camera independently outputs event data based on the principles outlined above, significantly reducing redundant data recording. Compared to traditional imaging devices, event cameras offer notable advantages in terms of low latency and low cost. Additionally, by utilizing a logarithmic operation mechanism to process light intensity signals, event cameras can effectively operate under extreme lighting conditions, demonstrating a high dynamic range. As a result, event cameras show great potential in observing complex, high-dynamic, and high-velocity targets.

\subsection{Leading-edge features extraction based on event-based spatiotemporal distribution monotonicity}
\noindent When observing high-dynamic, high-velocity targets, event cameras often exhibit a tailing effect due to various complex factors. This effect is characterized by the generation of a significant number of tailing events along the target's motion trajectory, which are triggered in close temporal and spatial proximity to those triggered by the high-speed target itself, as illustrated in Fig. 2. This close proximity leads to a shift in the centroid during the feature extraction process, which subsequently introduces errors in the calculation of the target's three-dimensional position. For high-speed moving targets, even small positional deviations within a brief time interval can result in significant inaccuracies in speed estimation.
\begin{figure}[H]
  \centering
  \includegraphics[width=0.48\textwidth,height=!]{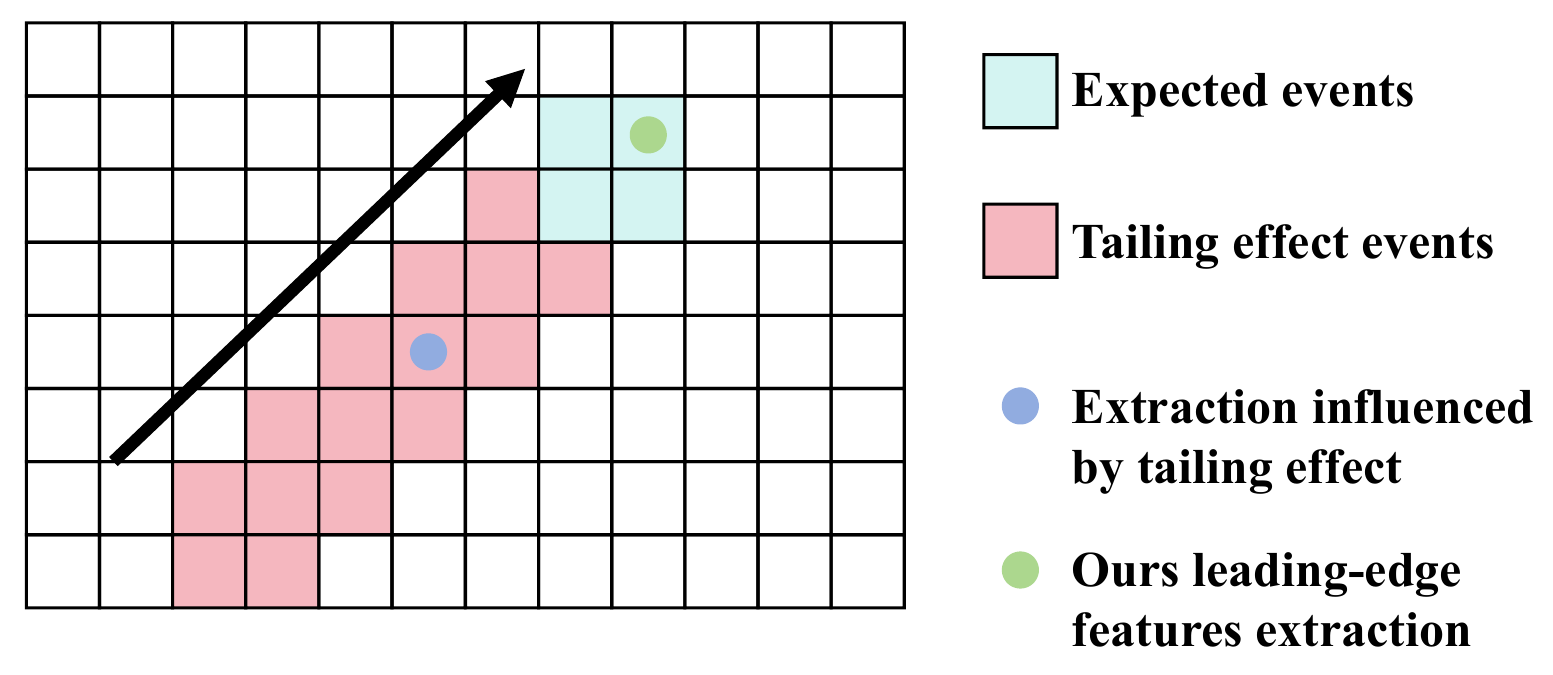}
  \caption{The impact of the event tailing effect of high-velocity targets on measurement. It can be observed that the tailing effect seriously affects the accurate extraction of the target position.}
\end{figure}
To address the challenge of accurately estimating the motion parameters of high-velocity targets due to the event tailing effect, this section proposes a method that utilizes the target's leading-edge features to estimate the motion parameters. Additionally, a target leading-edge features extraction algorithm based on the spatiotemporal distribution monotonicity of events is designed, as illustrated in Fig. 3. Considering that the distance between the target and the scattering origin increases monotonically over time, the approach proceeds as follows: First, for each view of the photogrammetric system, the events are projected onto a histogram based on distances from the origin. The distance in the histogram increases monotonically over time, and outliers are removed. The first event is set as the reference event. Next, based on the timestamps of the events, all events are traversed in increasing order. If the current event satisfies the condition in Eq. (3) relative to the reference event:
	\begin{eqnarray}
            \parallel (x_{c}-x_{0},y_{c}-y_{0})\parallel-\parallel (x_{r}-x_{0},y_{r}-y_{0})\parallel \textgreater 0.
    \end{eqnarray}
it is designated as the new reference event. Finally, the set of all reference events generated during this process forms the leading-edge features event set of the high-velocity fragment target. The formalization of this algorithm is shown in the Alg. 1.
\begin{algorithm}[H]
    \renewcommand{\algorithmicrequire}{\textbf{Input:}}
	\renewcommand{\algorithmicensure}{\textbf{Output:}}
	\caption{Extraction of target leading-edge features events}
    \label{power}
    \begin{algorithmic}[1]
        \REQUIRE Event set $A$;
	    \ENSURE Target leading-edge features event set $H$;
        \STATE $e_{r}=e_{1}$; $H=\varnothing$;
        \WHILE{c \textless $card(A)$}
        \IF {fit Eq. (3)}
            \STATE $e_{r}=e_{c}$; $c=c+1$; $H=H \cup e_{c}$;
        \ELSE \STATE {$c=c+1$};
        \ENDIF
        \ENDWHILE
        \STATE \textbf{return} $H$.
    \end{algorithmic}
\end{algorithm}
\begin{figure}[H]
  \centering
  \includegraphics[width=0.42\textwidth,height=!]{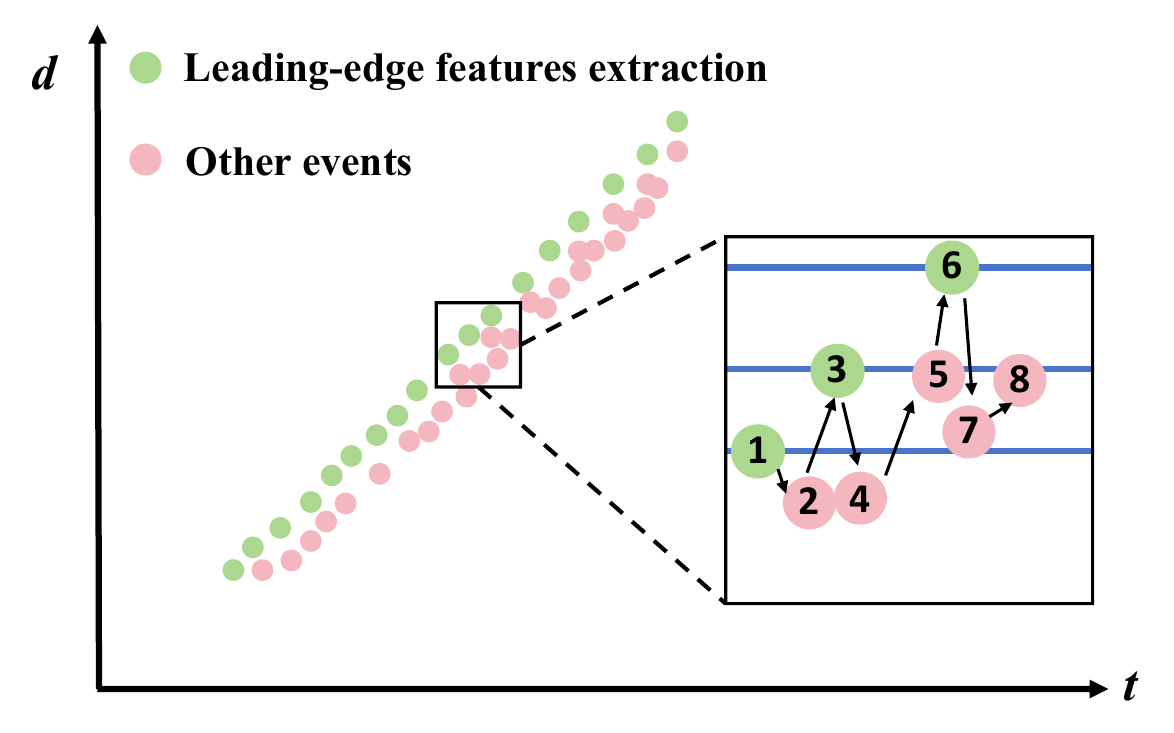}
  \caption{Our target leading-edge features extraction algorithm. The numbers in the figure represent the traversal order.}
\end{figure}
\subsection{Event-target trajectory association based on reprojection error}
Traditional multi-view geometry methods extract corresponding points from multiple views. By combining the intrinsic and extrinsic parameters of the camera system, corresponding-point-based intersection are used to determine the 3D spatial position of the target. However, events triggering in event cameras are asynchronous and can be random. When a moving target is close to the background brightness, it may fail to trigger an event. This complicates the concept of point correspondences for events. In high-dynamic or high velocity target photometry, asynchrony and randomness may cause event mismatches, leading to significant intersection errors. This reduces the robustness of the system. While stricter constraints on point correspondences (epipolar constraint, time windows etc.) can reduce errors, they also significantly decrease the number of events available for calculation. As a result, the accuracy and stability of the final solution are impacted.
\begin{figure}[H]
  \centering
  \includegraphics[width=0.48\textwidth,height=!]{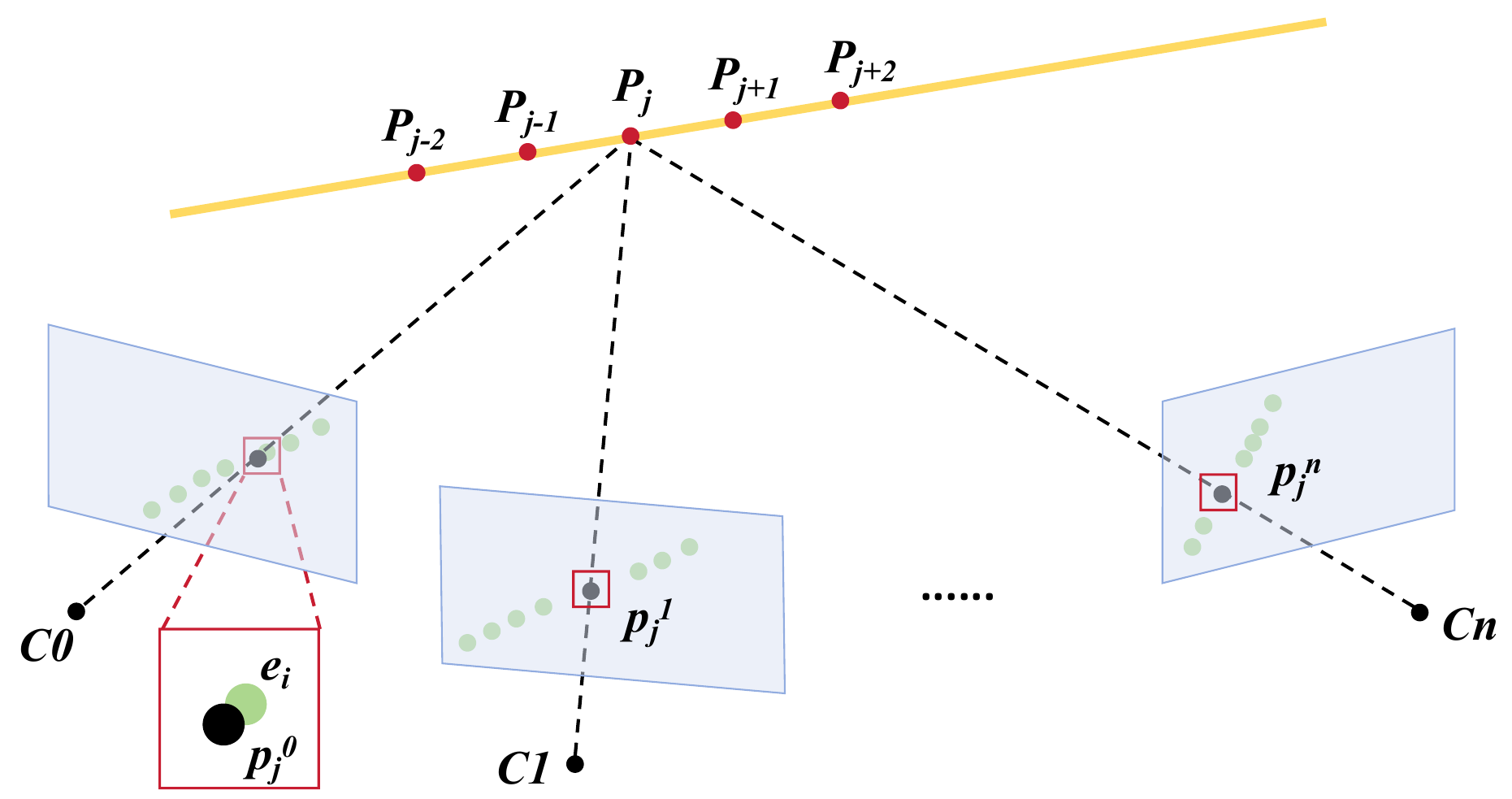}
  \caption{Reprojection error based matching for trajectory and events. After traversal, the projection point $p_{j}^0$ of $P_j$ is closest to event $e_i$ and less than the receiving threshold $\Omega$, so it is considered that $P_j$ matches $e_i$.}
\end{figure}

Therefore, this paper proposes a new measurement paradigm. It avoids the traditional intersection method based on point correspondences. Instead, it leverages the more robust trajectory line features in event cameras. This algorithm enables accurate measurements when a single event camera captures target motion, eliminating the need for simultaneous binocular capture. This significantly expanding the observable range of target motion. Specifically, the 3D scattering trajectory is first determined. Then, multi-view events are matched to the 3D points along the trajectory, establishing the relationship between the 3D coordinates and timestamps. The specific process is shown in Fig. 4.

First, the algorithm from section 3.2 is applied to extract the leading-edge features event sets $H_{1},H_{2},...,H_{n}$ for each view. Next, the target trajectories in each view are fitted using line fitting. By combining the intrinsic and extrinsic parameters of the calibrated stereo event camera system, the corresponding-line-based intersection method is used to obtain the target's scattering trajectory in 3D space. Then, starting from the initial search point $P_{j}(X_j,Y_j,Z_j,1)$, the search is performed along the 3D trajectory with a small step size. For each search point, the projection relationship in each view is governed by the pin-hole imaging model of the event camera, as described by Eq. (4)\cite{28}:
	\begin{eqnarray}
            \left [
            \begin{array}{cc}
                x_{j}^n\quad y_{j}^n\quad 1
            \end{array}
            \right ]^T=\bm{\mathit{\Pi_n}}
            \left [
            \begin{array}{cc}
                X_j\quad Y_j\quad Z_j\quad 1
            \end{array}
            \right ]^T
        \end{eqnarray}
where $\bm{\mathit{\Pi_n}}$ represents the projection matrix of the event camera $n$, which can be synthesized from the camera's intrinsic and extrinsic parameters using Eq. (5):
         \begin{eqnarray}
            \bm{\mathit{\Pi_n}}=\left [
            \begin{array}{ccc}
                f_{x}^n & 0 & o_x \\
                0  & f_{y}^n  & o_y \\
                0  & 0  & 1
            \end{array}
            \right ]
            \left [
            \begin{array}{cc}
                \bm{\mathit{R_n}} \lvert \bm{\mathit{T_n}}
            \end{array}
            \right ].
        \end{eqnarray}
where $(f_{x}^n,f_{y}^n),(o_x,o_y),\bm{\mathit{R_n}},\bm{\mathit{T_n}}$ are the focal length, principal point, rotation matrix, and translation vector of the event camera $n$ respectively. For each search point, reprojection is performed according to Eq. (4) and (5). If, for any event $e_i(x_i,y_i,t_i,p_i)$, the Euclidean distance between the event and the reprojection point is smaller than the smaller value between the reception threshold and the existing recorded value, the Euclidean distance is updated as the new recorded value for that event, as shown in Eq. (6).
	\begin{eqnarray}
            \begin{array}{ll}
             \Omega _{i}=\parallel (x_{j}^n-x_i,y_{j}^n-y_i) \parallel, \\
            if \parallel (x_{j}^n-x_i,y_{j}^n-y_i) \parallel \textless \min (\Omega,\Omega _{i})
            \end{array}
    \end{eqnarray}
where $\Omega$ represents the reception threshold, $\Omega _{i}$ and is the recorded value of the event $e_i$. It is important to note that the initial value of the recorded value is typically set to be greater than the reception threshold $\Omega$. The value of the threshold $\Omega$ is directly linked to the calibration reprojection accuracy. We set the threshold $\Omega$ to $2$pixel since the largest reprojection error in our experiment is $1.67$pixel.

After the above search procedure, the 3D target trajectory points are associated with each event. For any event, if its final recorded value $\Omega _{i}$ is smaller than the reception threshold $\Omega$, the association is considered successful. The event's timestamp $t_i$, along with the corresponding 3D target trajectory point $P_j(X_j,Y_j,Z_j,1)$, is then encoded as a photogrammetric data point $P_j^{'}(P_j,t_i)$. Otherwise, the association is considered unsuccessful, and the event is discarded. The overall algorithm flow is shown in the Alg. 2.
\begin{algorithm}[H]
    \renewcommand{\algorithmicrequire}{\textbf{Input:}}
	\renewcommand{\algorithmicensure}{\textbf{Output:}}
	\caption{Reproject error based matching for events and 3D target trace}
    \label{power}
    \begin{algorithmic}[2]
        \REQUIRE The set of search points on the target trajectory $\{P_1,P_2,...,\}$, multiple view projection matrices $\Pi_1,\Pi_2,...,$, the sets of target leading-edge features events from multiple views $H_1,H_2,...,$. the reception threshold $\Omega$.
	    \ENSURE Data set $\iota$;
        \STATE $H=H_1 \cup H_2 \cup H_3...$; $\iota=\varnothing$;
        \FOR{$P_j \in \{P_1,P_2,...,\}$}
            \FOR{$e_i \in H$}
                 \IF {fit Eq. (6)}
                 \STATE update $\Omega_i$; record the index of event $i$ and point $j$;
                 \ENDIF
            \ENDFOR
        \ENDFOR
        \STATE update $\iota$ according to final recorded indexes of events and points;
        \STATE \textbf{return} $\iota$.
    \end{algorithmic}
\end{algorithm}

\subsection{Target motion measurements based on discrete data.}
\noindent Using the method described in section 3.3, the correspondences between the 3D position and timestamp of a high-dynamic, high-velocity target, along with their evolution, is established. At this point, all data from different event cameras collectively form the set:
	\begin{eqnarray}
             \iota=\{ P_j^{'} \lvert \exists n, i; (\Omega_i \textless \Omega \land \parallel \Pi_nP_j^{T}-[x_i,y_i,1]\parallel=\Omega_i) \}.
    \end{eqnarray}
    
For high-velocity targets, directly calculating velocity based on the adjacent coordinates often introduces significant transmission errors due to the small time intervals. This results in poor stability of the measurements. An alternative approach is to select a larger interval for measurement. While this can improve result stability, there is insufficient theoretical guidance on how to appropriately choose the interval, which leads to lower reliability of the final results. To address this, this section uses high-velocity fragments as an example and applies the theoretical formula for fragment dispersal mechanics to fit the photogrammetric data point set. The theoretical formula for the decay of fragment velocity over time is as follows\cite{29}:
	\begin{eqnarray}
         v=\frac{v_0}{1+kv_0t}
    \end{eqnarray}
where $v$ represents the current velocity of the fragment, and $v_0$ represents the initial velocity of the fragment. $k$ is a parameter that characterizes the fragment's ability to maintain its velocity during flight, referred to as the fragment decay coefficient\cite{30}:
	\begin{eqnarray}
         k=\frac{c_x \rho s}{2M}
    \end{eqnarray}
where $M$ represents the mass of the fragment, $\rho$ is the air density, $s$ is the frontal area of the fragment, and $c_x$ is the drag coefficient. By integrating both sides of the equation with respect to time, the theoretical formula for the fragment's displacement $D$ over time is obtained as follows:
	\begin{eqnarray}
         D=\mathrm{\int} v \mathrm{d}t=\frac{1}{k} \ln \lvert 1+kv_0t \lvert+C
    \end{eqnarray}
where $C$ is a constant. By fitting the set $\iota$ using the above formula, the initial velocity of the fragment $v_0$ and its evolution curve can be directly obtained.

\section{Experiments}

\subsection{Experimental environment and instrument configuration}
To verify the effectiveness of the proposed method, field experiments are conducted on high-dynamic, high-velocity fragments from a light gas gun. The setup of the photogrammetric system is shown in Fig. 5. Two event cameras, synchronized with acquisition equipment, are used along with $400$W blue light lamps to illuminate the field of view and monitor the $1$m dispersion region of the light gas gun. The event cameras used in the experiment have a minimum time resolution of $1\upmu\mathrm{s}$ and a spatial resolution of $1280 \times 720$pixel. The length of the light gas gun barrel is $1.5$m, with a gas chamber volume of approximately $50$L. The fragment caliber is $8$mm, with a mass of approximately $5$g, and the air pressure is $5$Mpa. In the experiment, a chessboard calibration plate is moved across the common field of view, and the Zhang's calibration method\cite{31} is used to calibrate the intrinsic and extrinsic parameters of the event camera system. Some of the experimental equipment is shown in Fig. 6.
\begin{figure}[H]
  \centering
  \includegraphics[width=0.48\textwidth,height=!]{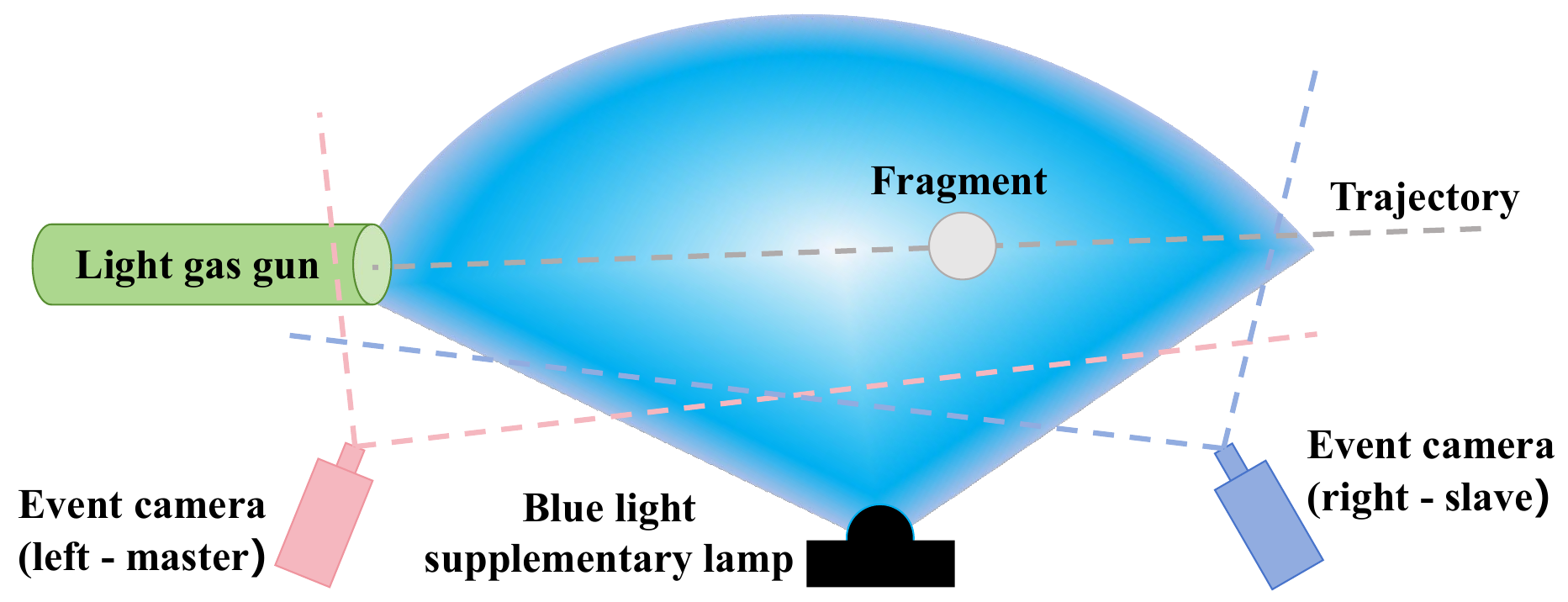}
  \caption{Diagram of the overall experimental setup. The blue light (450-490nm) enhances contrast and performs better in penetrating smoke and dust, making it particularly effective for high dynamic and high-speed fragment tracking.}
\end{figure}
\begin{figure}[H]
  \centering
  \includegraphics[width=0.48\textwidth,height=!]{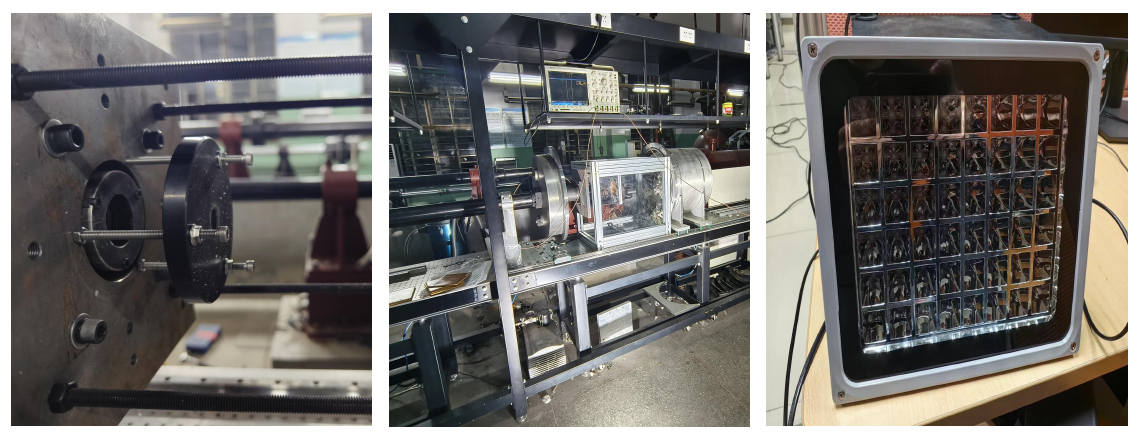}
  \caption{Diagram of selected experimental equipment: left (light gas gun - near), middle (light gas gun - far), right (blue light supplementary lamp).}
\end{figure}
\begin{figure}[H]
  \centering
  \includegraphics[width=0.48\textwidth,height=!]{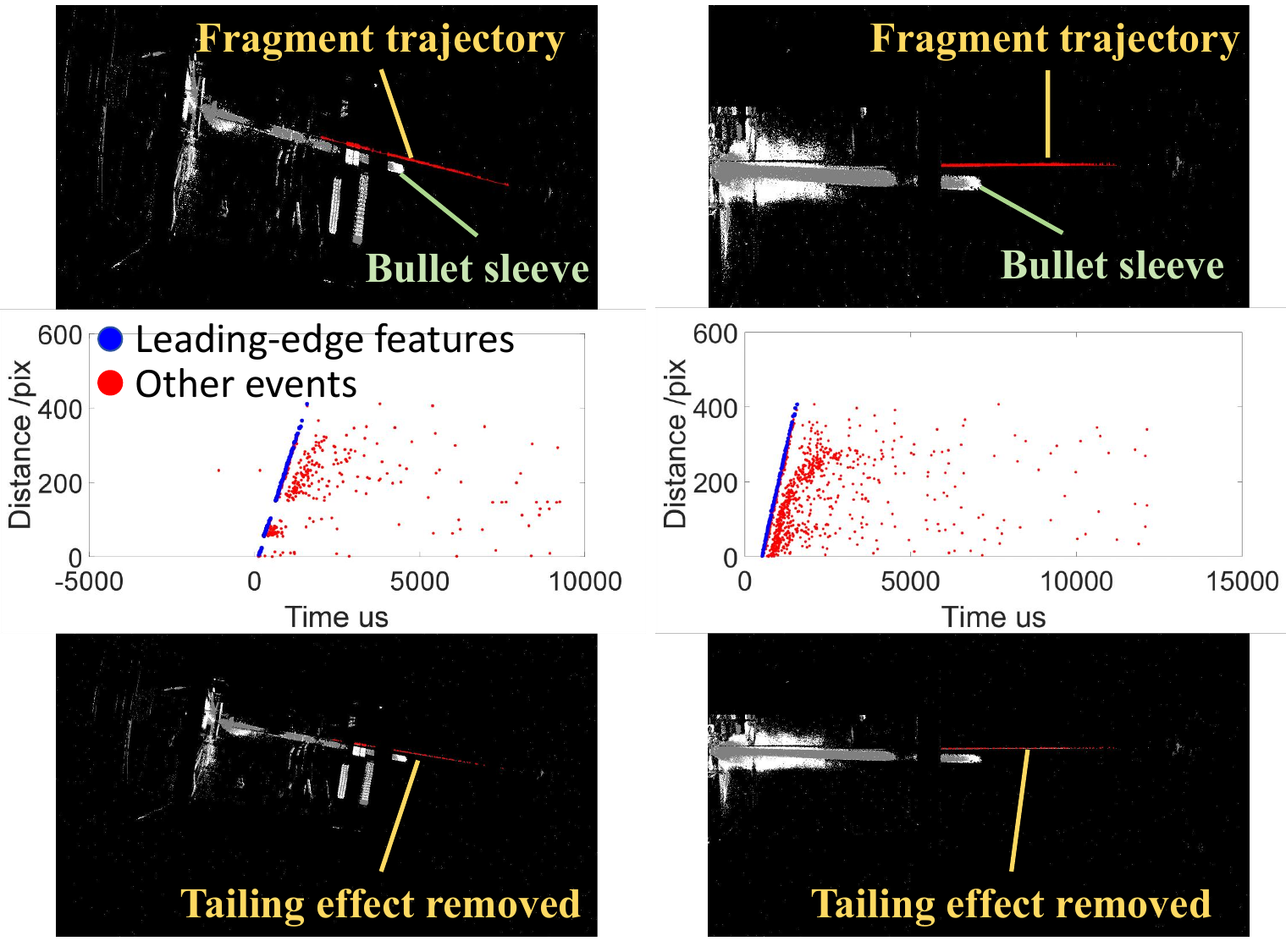}
  \caption{High-velocity target leading-edge features extraction performance: elimination of tailing effects (left: master event camera) (right: slave event camera). The "Distance" represents the Euclidean pixel distance from the first event, the horizontal axis represents the timestamp.}
\end{figure}

\subsection{Extraction of high-velocity fragment leading-edge features events}
The algorithm described in section 3.2 is used to extract leading-edge features events from all events on the target trajectory, effectively eliminating the interference of the tailing effect on the measurement of the high-velocity fragment target's motion parameters. Fig. 7 shows the results before and after the algorithm processing, along with intermediate outputs. By accumulating asynchronous events into event frames, the algorithm's effectiveness is visually demonstrated. Simultaneously, all events near the trajectory are projected into a 2D histogram, with the vertical axis representing the distance from the starting point and the horizontal axis representing time. The events in this histogram include noise events, fragment target events, and fragment tailing effect events. Blue points represent leading-edge features events of a fragment, while red points represent other events. Experimental results show that the leading-edge features events of the fragment target can be effectively separated from all events near the trajectory, enabling accurate measurement of the target's motion parameters.

\subsection{Measurement of fragment motion parameters}
\begin{figure}[H]
  \centering
  \includegraphics[width=0.45\textwidth,height=!]{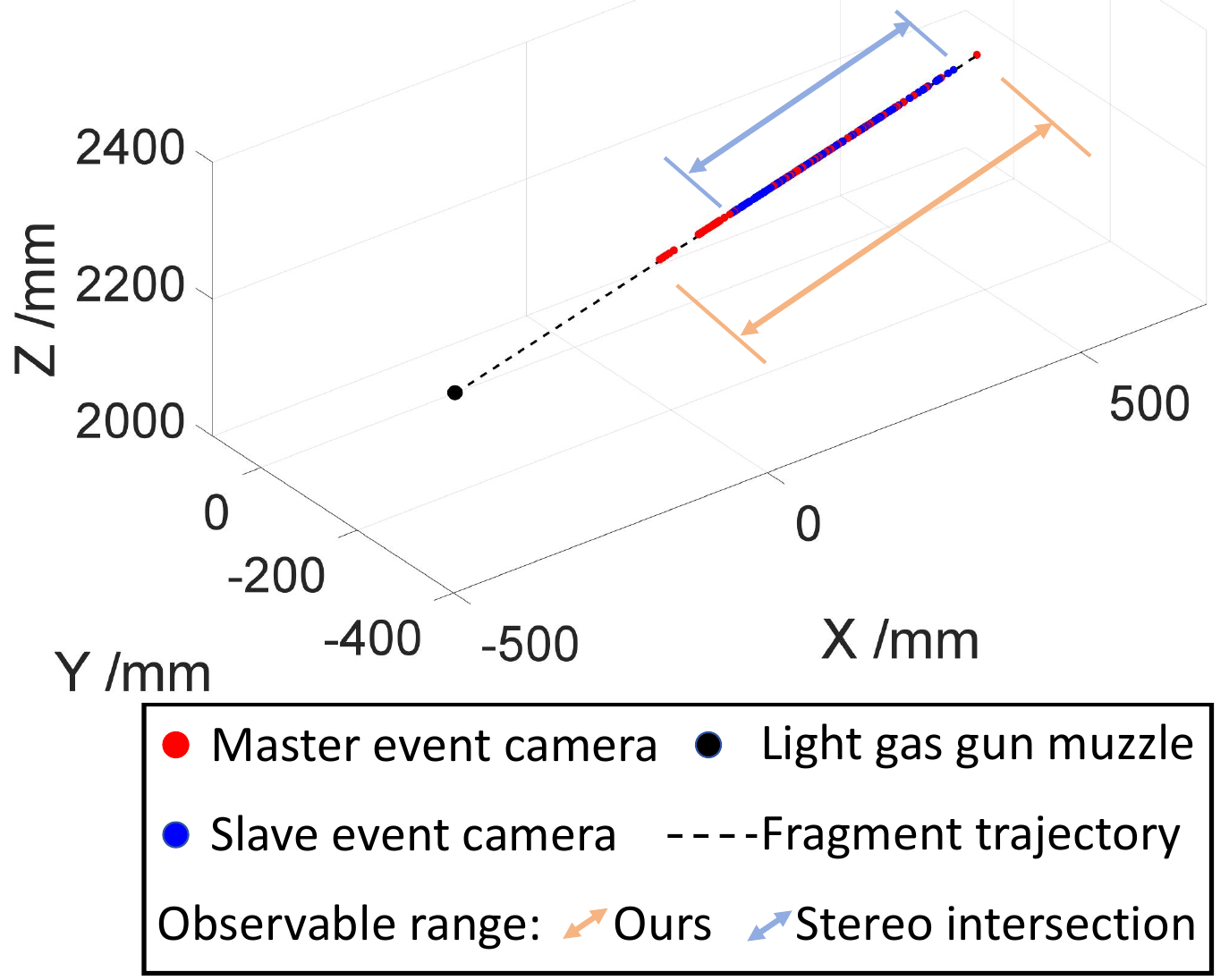}
  \caption{Light gas gun fragment 3D trajectory measurement results. Red dots represent the data from the main event camera, while blue dots represent the data from the secondary event camera. Our method provides more data than traditional intersection methods.}
\end{figure}
The algorithm described in section 3.3 is used to retrieve the corresponding three-dimensional spatial points for each event, with a retrieval step size set to $0.01$mm. The position information of the three-dimensional spatial points is combined with the timestamp information of the events. The result is shown in Fig. 8, where black points represent the light gas gun muzzle position, red points represent the data provided by the main camera, and blue points represent the data provided by the other camera.

Furthermore, the theoretical formula for the fragment's scattering distance over time, as described in section 3.4, is used to fit the data provided by the main and secondary cameras. The Trust Region algorithm is employed for fitting optimization. The green curve in Fig. 9 shows the fitting results. Based on the fitting parameters, the initial velocity of the fragment target is estimated to be $385.6$m/s.
\begin{figure}[H]
  \centering
  \includegraphics[width=0.48\textwidth,height=!]{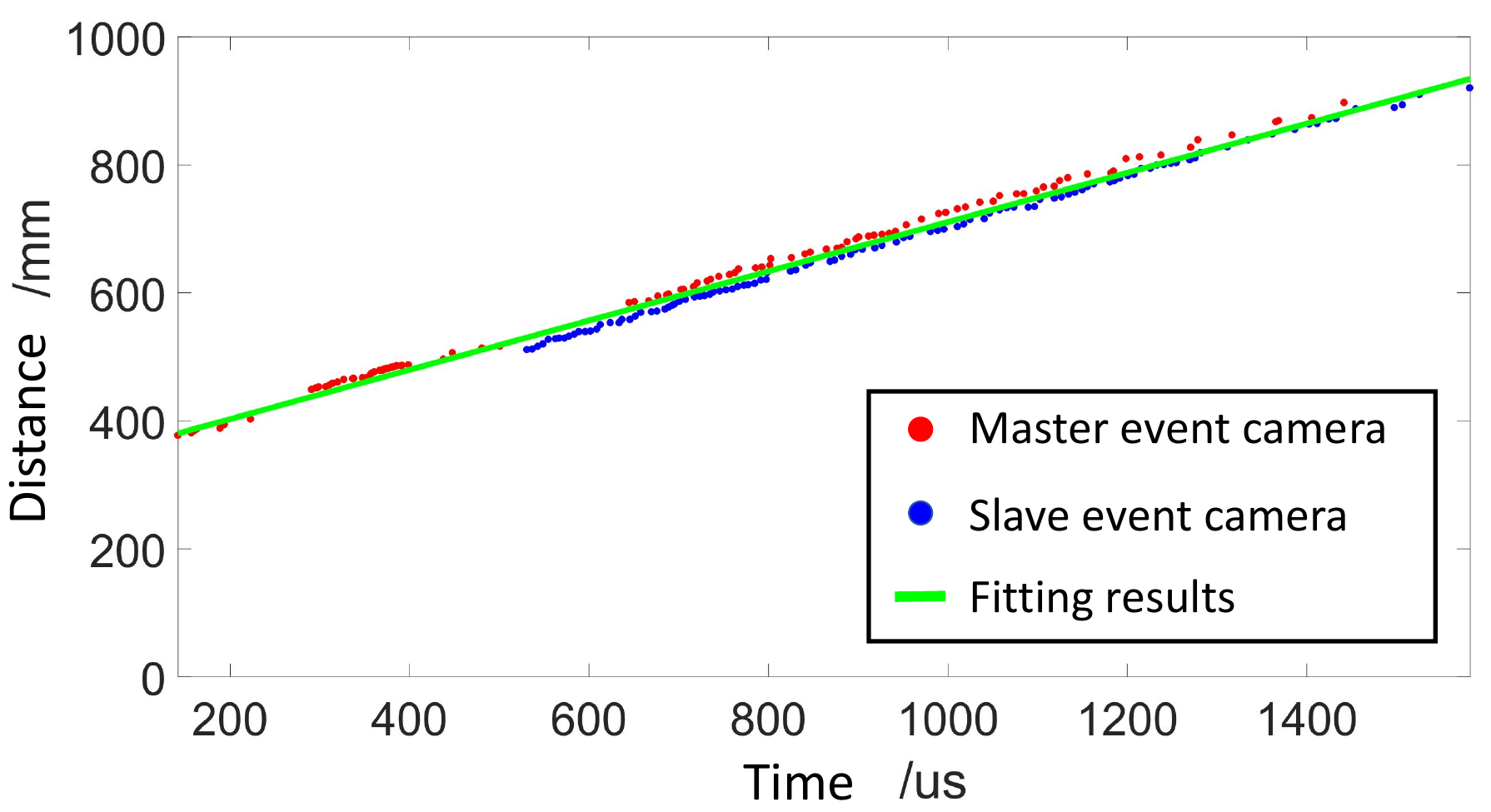}
  \caption{Fragment scattering distance over time measurement results. Red dots: the data from the main event camera, blue dots: the data from the secondary event camera, green line: the fitting results.}
\end{figure}

To validate the method's accuracy, the results from the electromagnetic speedometer are used for comparison. The air cannon work theory is also applied to predict the initial velocity as a reference. However, this prediction is influenced by practical factors. The relationship between the gas chamber volume, initial pressure, barrel length, and fragment initial velocity can be derived based on the work done by the gas during the firing process \cite{32}:
	\begin{eqnarray}
         \varphi E=\mathrm{\int} _0 ^{l_0} Q_0 \frac{V^{\gamma} s}{(V+sl)^\gamma} \mathrm{d}l
    \end{eqnarray}
where the kinetic energy of the fragment at the muzzle is represented by $E$, the barrel length by $l_0$, the initial pressure in the gas chamber by $Q_0$, the gas chamber volume by $V$, the secondary work coefficient by $\varphi$, the gas polytropic index by $\gamma$, the bore cross-sectional area by $s$, and the fragment's travel distance by $l$. After predicting the initial velocity, a curve is plotted using the formula for fragment velocity decay with displacement for comparison\cite{33}:
	\begin{eqnarray}
         v=v_0\exp (-kD).
    \end{eqnarray}
Based on the experimental conditions, the values of the above parameters are as follows: $l_0=1.5\mathrm{m}, Q_0=5\mathrm{Mpa}, V=50\mathrm{L},\varphi=1.05,\gamma=1.2,s=5.03×10^{-5}\mathrm{m}^2$. This leads to an estimated initial velocity of the fragment as $378.8$m/s. Table 1 compares the fragment velocity obtained through the event camera system's optical method, the electromagnetic speedometer results, and the theoretical predictions.
\begin{table}[H]
  \centering
    \caption{Fragment initial velocity measurement results.}
  \begin{tabular}{c c c}
  \hline  
   & initial velocity & deviation \\
  \hline
  \textbf{Ours} & \textbf{385.6$\mathrm{m/s}$} & - \\
  \multirow{2}{*}{Electro metric} & \multirow{2}{*}{369.1$\mathrm{m/s}$} & $+16.5$(absolute) \\ 
  & & 4.47\%(relative) \\
  \multirow{2}{*}{Theory} & \multirow{2}{*}{378.8$\mathrm{m/s}$} & $+6.80$(absolute) \\ 
  & & 1.80\%(relative) \\
  \hline
  \end{tabular}
\end{table}
Compared to traditional event-based stereo intersection methods, our approach offers more data. The traditional method, with a 20$\upmu\mathrm{s}$ time window and a 2pixel epipolar constraint, provides only 6 points. In contrast, our method yields a total of 213 points for measurements.

The experimental results show that the high-dynamic, high-velocity target measurement method based on multi-view event cameras proposed in this paper demonstrates good accuracy and stability. It can provide a large number of effective data for high-velocity targets with high time resolution (equivalent frame rate around $146,000$fps), showcasing strong advancement and practical application value.

\section{Conclusion}
\noindent This paper adopts a photogrammetric system based on event cameras and proposes a high-dynamic, high-velocity target measurement method handling event stream data. First, the monotonicity of the spatiotemporal distribution of events is utilized to extract the target's leading-edge features, eliminating the tailing effect that impacts motion measurements. Next, the reprojection error is used to establish a connection between events and the target's 3D trajectory points. Finally, a target velocity attenuation theoretical model is employed to fit the data obtained from the event stream, enabling the measurement of high-dynamic, high-velocity target motion parameters via ours multi-view data joint computation. 

Experimental validation, conducted on light-gas gun fragments, demonstrates the method's effectiveness. The comparison of initial velocity measurements with results from an electromagnetic speedometer and theoretical predictions reveals deviations of only $4.47\%$ and $1.80\%$, respectively. These findings highlight the significant advantages of the proposed method, including continuous observation, high time resolution, high accuracy, and a wide dynamic range, compared to traditional techniques. Furthermore, this method offers a promising low-cost approach for high-dynamics mechanical testing, with broad applicability in industrial fields.

  \vspace{1em}
  \begin{fontsize}{8}{0.8}
\noindent
\textit{This work was supported by the National Natural Science Foundation of China (Grant No. 12372189) and the Hunan Provincial Natural Science Foundation for Excellent Young Scholars (Grant No. 2023JJ20045).}
\end{fontsize}

\renewcommand{\mkbibbrackets}[1]{#1.}

\end{multicols}


\makeentitle

\end{document}